\documentclass[11pt]{article}
\usepackage{coling2018}
\usepackage{times}
\usepackage{url}
\usepackage{latexsym}
\usepackage{CJKutf8}
\usepackage{graphicx}
\usepackage{caption}
\usepackage{subcaption}

\title{Multi-Layer Ensembling Techniques for Multilingual Intent Classification}

\author{Charles Costello, Ruixi Lin, Vishwas Mruthyunjaya, Bettina Bolla, Charles Jankowski \\CloudMinds Technology Inc\\Santa Clara, CA, USA\\ \{charlie.costello, ruixi.lin, vish.m, bettina.bolla, charles.jankowski\}@cloudminds.com}
\date{}

\begin{document}
\maketitle
\begin{abstract}
In this paper we determine how multi-layer ensembling improves performance on multilingual intent classification. We develop a novel multi-layer ensembling approach that ensembles both different model initializations and different model architectures. We also introduce a new banking domain dataset and compare results against the standard ATIS dataset and the Chinese SMP2017 dataset to determine ensembling performance in multilingual and multi-domain contexts. We run ensemble experiments across all three datasets, and conclude that ensembling provides significant performance increases, and that multi-layer ensembling is a no-risk way to improve performance on intent classification. We also find that a diverse ensemble of simple models can reach perform comparable to much more sophisticated state-of-the-art models. Our best $F_1$ scores on ATIS, Banking, and SMP are 97.54\%, 91.79\%, and 93.55\% respectively, which compare well with the state-of-the-art on ATIS and best submission to the SMP2017 competition. The total ensembling performance increases we achieve are 0.23\%, 1.96\%, and 4.04\% $F_1$ respectively.
\end{abstract}

%
% The following footnote without marker is needed for the camera-ready
% version of the paper.
% Comment out the instructions (first text) and uncomment the 8 lines
% under "final paper" for your variant of English.
% 
\blfootnote{
    %
    % for review submission
    %
    \hspace{-0.65cm}  % space normally used by the marker
    Place licence statement here for the camera-ready version.
    % 
    % % final paper: en-us version 
    %
    % \hspace{-0.65cm}  % space normally used by the marker
    % This work is licensed under a Creative Commons 
    % Attribution 4.0 International License.
    % License details:
    % \url{http://creativecommons.org/licenses/by/4.0/}
}

\section{Introduction} \label{Introduction}

In this work we determine how beneficial multi-layer ensembling is for intent classification in multilingual and multi-domain contexts. Our hypothesis is that multi-layer ensembles can provide significant performance increases compared to their constituent models. Additionally, we test whether ensembles of simple models can compete with more complex models, and how well ensembled models generalize across different languages and domains. 

Our contributions in this work are threefold: 
\begin{itemize}
\item We present an exhaustive performance comparison of different CNN and RNN multi-layer ensembles across three different datasets on the intent classification task. We run experiments on 12 individual models and 4083 separate ensembles.
\item We develop a novel multi-layer ensembling approach. As opposed to focusing on either ensembling different random initializations or different model architectures, as most ensembling approaches do, we combine both methods in order to fully explore the possible performance improvements of ensembling. 
\item We make a novel comparison of ensemble results across multiple domains and languages. This multilingual and mulit-domain analysis allows us to discuss the generalizability of ensembling techniques, as well as their potential performance gains in different contexts.
\end{itemize}

Our paper is outlined in the following manner. In section \ref{Related Work} we discuss other work related to our own. In section \ref{Data} we describe and compare the three datasets.  In section \ref{Technical Background} we discuss the technical background for our work, including the intent classification task, the individual CNN and RNN models, and the multi-layer ensembling approach.  In section \ref{Experiments and Results} we present a detailed description of our experiments and an analysis of the results. Finally, in section \ref{Conclusion} we draw general conclusions from our work. 

\section{Related Work} \label{Related Work}

As an essential component of NLP systems, intent classification has been thoroughly researched. Although approaches such as SVMs \cite{Sheikh2016}, neural bag-of-words \cite{Purohit2015}, and knowledge guided patterns \cite{Bhargava2013} have proven successful, deep learning has been shown to be a more effective algorithm to solve classification tasks in natural language processing \cite{Mikolov2010,Shi2016}.

The two major deep learning variants used for intent classification are Convolutional Neural Networks (CNNs) and Recurrent Neural Networks (RNNs). Although CNNs are heavily used in computer vision tasks, they have also been adapted and proven successful on NLP tasks \cite{Collobert2008,Shi2016}. RNN approaches have also worked very well for intent classification \cite{Ravuri2015,Socher2012}. In popular cases, intent classification resides alongside slot filling as a jointly modeled NLP task using deep learning algorithms \cite{Bhardwaj2017,Hashemi2016}. For instance, the CNN based triangular CRF architecture exploits the dependency between intent and slots \cite{Xu2013}, whilst the attention-based RNN model includes attention vectors similar to those used in attention-based encoder-decoder models \cite{Liu2016}. Another such joint classification model is an RNN architecture that uses GRU representations to identify intents and slots \cite{Zhang2016}. 

Although CNNs and RNNs have shown impressive performance on many NLP tasks, individual models are still susceptible to overfitting and local optima. To combat this, many researchers have explored ensembling multiple models to obtain higher quality inference \cite{Rokach2010}. These methods have been used widely in NLP research \cite{Jin2017}, and \newcite{Purohit2015} used single-layer ensembling in their work on intent classification.

\section{Data} \label{Data}

\subsection{ATIS}

The ATIS (Airline Travel Information System) dataset was collected by DARPA in the early 90's \cite{Price1990,Tur2010}. It is one of the most frequently used dataset for SLU (spoken language understanding) research. It was based on spoken queries regarding flight related information. An example utterance is: ``I'd like to fly from Boston to San Francisco.'' The dataset exists in both spoken and text form; we use the text form here.

The utterances are represented by semantic frames, where each sentence has a single or multiple intents and slots filled with phrases. Labels are encoded with IOB (Inside Outside Beginning) representation \cite{Sang1999}. There are 17 single intents and 8 multiple intent combinations. The distribution is biased as the most frequent intent, Flight, represents about 70\% of the traffic. The ATIS train set contains 4,978 sentences. The test set contains 893 sentences. 

\subsection{Banking}

For a particular banking-related application, we wanted to get domain specific data. Since most banking data is proprietary, we had to collect our own. Our proprietary banking dataset was collected as an initial in-house pilot study of 360 usable written utterances and increased to a total of 5,358 usable (clean) utterances by utilizing Mechanical Turk. An example utterance is: ``I would like to open a checking account.'' Data collection was based on 12 prompts representing different slot/intent combinations.

The utterances were tokenized and annotated following the IOB representation used in the ATIS dataset, with single or multiple intents. There are 9 single intents, including the Null Intent, and 10 dual intent combinations. The distribution is somewhat skewed towards the Find\_Out intent, but is much more evenly distributed than ATIS. Our banking train set contains  4,286 sentences. The test set contains 1,072 sentences. 

\subsection{SMP}

The SMP2017ECDT (SMP) dataset is a Chinese domain classification dataset consisting of textual queries recorded from human-computer dialogues \cite{Zhang2017}. We divide the datasets into a train and a test set of 3,069 and 667 samples respectively. The first step for many QA systems is to categorize a user intent into a specific domain, and we consider these domains as the first level user intents, for instance, \begin{CJK*}{UTF8}{gbsn}``结婚了吗''\end{CJK*} (Are you married) is labeled as Chit-chat, which implies that the user intent is related to chit-chatting. The domains cover 31 categories, including Chit-chat and 30 task-oriented categories. The SMP dataset is skewed towards Chit-chat with around 20\% of data in it, and the rest of the 30 categories are more evenly distributed. Since the Chinese data given is not tokenized, we used the Jieba tokenizer\footnote{https://github.com/fxsjy/jieba} to tokenize the sentences.

\section{Technical Background} \label{Technical Background}

\subsection{Task}

Intent classification is the task of correctly labeling a natural language utterance from a predetermined set of intents. We can treat it as a multiclass classification task, and train a discriminative machine learning model to output a predicted classification for a given utterance. In cases where an utterance has multiple intents, we concatenate the intents and treat the result as a single, distinct intent. This allows us to treat the task as single-label, even when multiple intents are used.  We use CNN and RNN models to perform this statistical inference. These models are trained on labeled utterances using backpropagation and gradient descent. At inference time, our models compute the probability distribution over all intent classes for a given utterance conditioned on the trained weights. This probability for each intent label is calculated by taking the softmax of the model outputs, and the predicted class is then simply chosen by taking the argmax of the distribution. 

\subsection{CNNs} \label{cnnsection}

Convolutional Neural Networks (CNNs) are widely used for sentence classification tasks \cite{LeCun1995}. CNNs have proven to be effective for NLP tasks like sentence classification \cite{Kim2014}. Besides a baseline of a simple CNN with four interleaved layers of convolution and pooling, we developed three CNN variants, including a character-level model \cite{Santos2014}, CharCNN, an attention based \cite{Neumann2017} bidirectional model \cite{Vu2016}, ABiCNN, and a combination of both variants which is a character-level attentive bidirectional model, ACharBiCNN. We pad samples of variable lengths to the maximal sample length of all samples in a dataset in order to do mini-batch training. There are 50 and 100 convolution filters for the first and second convolution respectively.

The character approach \cite{Santos2014} learns a word representation from the characters that make up a word in a character embedding layer. The output of the character embedding layer will be a mini batch of sentences embedded with the combined vector, which is followed by the normal CNN convolution and pooling layers.

We also apply a bidirectional CNN structure \cite{Vu2016} and an attention mechanism \cite{Neumann2017}. Bidirectional NNs have shown better performances than unidirectional NNs in language modeling and machine translation \cite{Vu2016,Bahdanau2015}. Attention based neural models have been seen successfully used in pattern recognition \cite{OIshausen1993}, NLP \cite{Yin2016}, and computer vision tasks \cite{Sun2003}. Our attentive bidirectional CNN model is built as follows. A bidirectional CNN takes inputs from a positive and a negative time direction, so that the output gets information from both past and future states. A combined hidden layer output can thus be obtained by a weighted sum of the individual hidden layer outputs, activated by a non-linear activation function. Then attention mechanism is applied, which helps attend to important parts of a feature map by learning weights of each part. The attention based bidirectional hidden layer output is then input to a fully-connected layer, followed by a softmax layer. Combining the character embedding approach and the attention mechanism, we obtained the character-level attentive bidirectional CNN model.

\subsection{RNNs}

RNNs are deep neural networks designed to process sequential data. They have proven to be very effective for many natural language processing tasks, including language modeling \cite{Mikolov2010} and text generation \cite{Sutskever2011}. RNNs are distinguished from traditional deep neural networks by the use of a cell architecture that is sequentially unrolled for a specified number of timesteps. This allows the network to communicate through multiple timesteps, effectively remembering past information. However, it was noted by \newcite{Bengio1994} that RNNs suffer from the vanishing gradient problem when the number of timesteps exceed a small number, prohibiting the network from remembering long-term information. To correct this problem, many RNN architectures were suggested that would allow for longer-term memory. Two of the most popular, which we use in this work, are the Long Short-Term Memory (LSTM) cell \cite{Hochreiter1997}, and the Gated Recurrent Unit (GRU) cell \cite{Cho2014}. 

When developing RNN models for intent classification, we looked at many architectural considerations, including using a LSTM or GRU cell architecture, using character embeddings as described in Section \ref{cnnsection} \cite{Santos2014}, and using a bidirectional architecture \cite{Schuster1997}. We then created a model for each combination, resulting in 8 total models. Each model has 512 hidden units and a word embeddings size of 300. Models with character embeddings have a character embedding size of 300. Word and character embeddings were randomly initialized and trained along with the model's other weights, without the use of any pretrained vectors. We trained for 100 epochs with a batch size of 32 and a dropout rate of 0.5.

\subsection{Multi-layer Ensembling}

\begin{figure}[t]
  \centering
  \includegraphics[width=16cm]{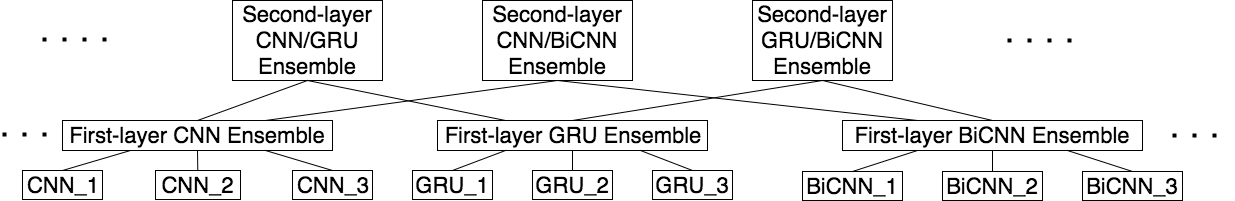}
  \caption{Multi-layer Ensemble Approach}
  \label{Multi-layer Ensemble Approach}
\end{figure}

We ran experiments on numerous multi-layer ensembles of our CNN and RNN models, as well as the individual models themselves, on all three datasets. We evaluated the ensembles and individual models on each test set, and used the unweighted $F_1$ score as our metric. When ensembling models, we use a majority vote with confidence approach \cite{Jin2017}. For each utterance, we first see if a majority of the models agree upon the prediction. If so, we choose that prediction. If not, we compare the confidence of each model's prediction as measured by the softmax score, and take the highest score. We chose majority vote with confidence as our ensembling method because it is a very simple and straightforward technique, and the goal of our work is to show how even simple ensembling can improve performance greatly.

Figure \ref{Multi-layer Ensemble Approach} shows our multi-layer ensemble. We can see that is comprised of distinct first-layer and second-layer ensembles that are arranged in a hierarchical structure. The first layer ensembles different random initializations of each model architecture (e.g. CNN\_1, CNN\_2, and CNN\_3). The second layer then ensembles the ensembled version of each model architectures (e.g. CNN ensemble and GRU ensemble). This allows the final ensembles to learn from both different random initializations and different model architectures.

More concretely, to obtain our first-layer ensembles, we first train each model three times with the same hyperparameters but different random initializations for the weights. We then use majority vote with confidence to find the ensemble predictions, and finally calculate the $F_1$ score. In Tables \ref{CNN First-layer Ensembles} and \ref{RNN First-layer Ensembles}, we show the three individual model $F_1$ scores (columns ``1'', ``2'', ``3''), as well as their ensemble $F_1$ (column ``En''). To obtain the second-layer ensembles, we generate all possible combinations of the first-layer ensembles that are of size two or more. As we have 12 first-layer ensembles, this gives us second-layer ensembles with $\{2, 3 \ldots 12\}$ number of first-layer ensembles. This yields $2^{12} - 12 - 1 = 4083$ distinct second-layer ensembles. We again use majority vote with confidence to evaluate the second-layer ensembles. Tables \ref{CNN Second-layer Ensembles} and \ref{Best Second-layer Ensembles} compare $F_1$ scores of different second-layer ensembles.

\section{Experiments and Results} \label{Experiments and Results}

\subsection{CNN Ensembles}

\begin{table}[t]
\centering
\begin{subtable}[t]{.475\linewidth}    
\begin{tabular}{ l | c | c | c | c }
 Model & 1 & 2 & 3 & En \\ 
 \hline
 CNN & 95.41 & 95.52 & 95.63 & 95.97 \\
 CharCNN & 91.94 & 92.50 & 93.39 & 94.40 \\
 ABiCNN & 96.64 & 96.75 & 97.09 & 97.20 \\
 ACharBiCNN & 93.73 & 93.84 & 94.29 & 95.07 \\
\end{tabular}
\caption{ATIS}
\label{CNN ATIS First-layer Ensembles}
\end{subtable}
\quad
\begin{subtable}[t]{.475\linewidth}    
\begin{tabular}{ l | c | c | c | c }
 Model & 1 & 2 & 3 & En \\ 
 \hline
 CNN & 87.22 & 87.59 & 88.90& 89.27 \\
 CharCNN & 87.59 & 87.69 & 88.53 & 89.18 \\
 ABiCNN & 88.99 & 89.37 & 89.83 & 90.11 \\
 ACharBiCNN & 88.25 & 88.71 & 89.09 & 90.11 \\
\end{tabular}
\caption{Banking}
\label{CNN Banking First-layer Ensembles}
\end{subtable}
\begin{subtable}[t]{.475\linewidth}    
\begin{tabular}{ l | c | c | c | c }
 Model & 1 & 2 & 3 & En \\ 
 \hline
 CNN & 85.46 & 85.61 & 86.06 & 87.26 \\
 CharCNN & 83.36 & 83.81 & 84.11 & 86.51 \\
 ABiCNN & 86.36 & 86.51 & 86.96 & 88.16 \\
 ACharBiCNN & 84.41 & 85.01 & 85.61 & 89.06 \\
\end{tabular}
\caption{SMP}
\label{CNN SMP First-layer Ensembles}
\end{subtable}
\caption{CNN First-layer Ensembles $F_1$ (\%)}
\label{CNN First-layer Ensembles}
\end{table}

\begin{table}[t]
\centering
\begin{subtable}[t]{.6\linewidth}    
\begin{tabular}{ l | c | c | c | c }
 Model & ATIS & Banking & SMP & Avg Inc \\
 \hline
CNN/CharCNN & 95.97 & 89.46 & 87.71 & 1.14 \\
CNN/ABiCNN &  96.70 & 90.49 & 89.81 & 2.10 \\
CNN/ACharBiCNN & 95.97 & 90.11 & 89.51 & 1.48 \\
CharCNN/ABiCNN & 96.19 & 90.58 & 88.76 & 1.70 \\
\end{tabular}\\
\caption{Interesting Two-model Ensembles}
\label{CNN Interesting Ensembles}
\end{subtable}
\quad
\begin{subtable}[t]{.35\linewidth}    
\begin{tabular}{ l | c | c }
 Dataset & $F_1$ & Ensemble  \\ 
 \hline
      ATIS & 97.20 & \begin{tabular}{@{}c@{}} ABiCNN \end{tabular} \\
      \hline
      Banking & 90.86 & \begin{tabular}{@{}c@{}} CNN, \\ ABiCNN, \\ ACharBiCNN \end{tabular} \\
      \hline
      SMP & 89.96 & \begin{tabular}{@{}c@{}} CNN, \\ CharCNN, \\ ABiCNN \end{tabular}
\end{tabular}
\caption{Best Ensembles}
\label{CNN Best Ensembles}
\end{subtable}
\caption{CNN Second-layer Ensembles $F_1$ (\%)}
\label{CNN Second-layer Ensembles}
\end{table}

\textbf{First-layer Ensembles}: The results of experiments on ensembling single CNN models are listed for ATIS, Banking, and SMP in Table \ref{CNN First-layer Ensembles}. The first thing that draws our attention is that first-layer ensembling improves the performances on all individual models  across all three datasets. Despite slight performance differences at each of the three times of training, the \(F_1\)scores of ensembles are higher than all of individual models. Comparing the ensemble \(F_1\) with the average \(F_1\) of all three individual models, there is a lowest 0.37\% improvement (ABiCNN on ATIS with individual scores: 96.64\%, 96.75\%, 97.09\%, ensemble score: 97.20\%) and a highest 4.05\% improvement (ACharBiCNN on SMP with individual scores: 84.41\%, 85.01\%, 85.61\%, ensemble score: 89.06\%). This shows that first-layer ensembles help to exploit the benefits of different random initializations, mitigating the risk of converging to local optima.

\textbf{Second-layer Ensembles}: In addition, with the first-layer ensemble results as input, the second-layer ensemble shows some interesting results when it comes to combining models of different structures. Out of the 11 second-layer ensembles of different combinations of the CNN models, we focused on comparing the ensemble of character embedding based single-model ensembles with the ensemble of simple CNNs, the ensemble of attention and bidirectional based single-model ensembles with the ensemble of simple CNNs, and the ensemble of character embedding based single-model ensembles with the ensemble of attention and bidirectional based ones. We did an exhaustive comparison across all 2-ensembles, 3-ensembles, and the all-ensemble, and found that the second-layer ensembles outperformed their component first-layer ensembles. We presented the most interesting 2-ensemble results in Table \ref{CNN Interesting Ensembles}. The average increase is calculated by averaging the increase for all three datasets, where the increase for each dataset is again given by the average of percentage increase of the ensemble (e.g. CNN/CharCNN) compared to its component single-model ensembles (e.g. CNN and CharCNN). 

We analyzed the idiosyncrasies of the best models across datasets. The best \(F_1\) on ATIS, Banking, and SMP (Table \ref{CNN Best Ensembles}) are 97.20\% with a minimal first-layer ABiCNN ensemble, 90.86\% with a minimal second-layer CNN/ABiCNN/ACharBiCNN ensemble, and 89.96\% with a minimal second-layer CNN/CharCNN/ABiCNN ensemble. It seems that the specific ensemble constituents differ from dataset to dataset, but there is a dominant  first-layer ensemble component ABiCNN appearing in all best ensembles across all three datasets, and it results from the fact that the attention structure and bidirectionality generally contribute additional contextual information to  learning, regardless of the language the data is in. From another aspect, despite of idiosyncratic components in the best models, a combination of different model structures, like combining simple structures with character embeddings, attention, and bidirectionality, will further help reduce False Positive rates as they compensate for one another on a probabilistic basis. 

\subsection{RNN Ensembles}

\begin{table}[t]
\centering
\begin{subtable}[t]{.475\linewidth}    
\begin{tabular}{ l | c | c | c | c }
 Model & 1 & 2 & 3 & En \\ 
 \hline
 GRU & 96.98 & 97.20 & 97.31 & 97.31 \\
 CharGRU & 95.07 & 95.30 & 95.41 & 95.63 \\
 BiGRU & 96.30 & 96.53 & 97.20 & 96.86 \\
 CharBiGRU & 94.18 & 94.29 & 94.40 & 94.62 \\
 LSTM & 96.19 & 96.53 & 96.86 & 96.98 \\
 CharLSTM & 95.07 & 95.97 & 96.86 & 96.53 \\
 BiLSTM & 96.75 & 96.75 & 96.86 & 97.09 \\
 CharBiLSTM & 93.95 & 94.51 & 94.74 & 94.74 \\
\end{tabular}
\caption{ATIS}
\label{RNN ATIS First-layer Ensembles}
\end{subtable}
\quad
\begin{subtable}[t]{.475\linewidth}    
\begin{tabular}{ l | c | c | c | c }
 Model & 1 & 2 & 3 & En \\ 
 \hline
 GRU & 87.69 & 87.78 & 88.53 & 88.15 \\
 CharGRU & 88.15 & 88.62 & 89.55 & 90.02 \\
 BiGRU & 86.94 & 87.69 & 88.15 & 88.15 \\
 CharBiGRU & 86.66 & 87.41 & 89.18 & 88.99 \\
 LSTM & 87.97 & 88.06 & 88.43 & 88.99 \\
 CharLSTM & 88.90 & 89.18 & 89.74 & 90.67 \\
 BiLSTM & 87.69 & 88.99 & 89.18 & 89.65 \\
 CharBiLSTM & 88.53 & 88.90 & 89.09 & 89.93 \\
\end{tabular}
\caption{Banking}
\label{RNN Banking First-layer Ensembles}
\end{subtable}
\begin{subtable}[t]{.475\linewidth}    
\begin{tabular}{ l | c | c | c | c }
 Model & 1 & 2 & 3 & En \\ 
 \hline
 GRU & 83.06 & 83.96 & 85.61 & 84.71 \\
 CharGRU & 86.81 & 88.31 & 89.36 & 89.96 \\
 BiGRU & 84.11 & 85.31 & 86.21 & 86.66 \\
 CharBiGRU & 85.16 & 87.11 & 87.41 & 88.46 \\
 LSTM & 84.41 & 85.76 & 85.91 & 87.26 \\
 CharLSTM & 88.01 & 88.31 & 89.51 & 91.30 \\
 BiLSTM & 83.81 & 84.86 & 86.06 & 86.36 \\
 CharBiLSTM & 87.56 & 88.16 & 88.76 & 89.51 \\
\end{tabular}
\caption{SMP}
\label{RNN SMP First-layer Ensembles}
\end{subtable}
\caption{RNN First-layer Ensembles $F_1$ (\%)}
\label{RNN First-layer Ensembles}
\end{table}

\begin{table}[t]
\centering
\begin{subtable}[t]{.475\linewidth}    
\begin{tabular}{ l | c | c }
 Dataset & $F_1$ & Ensemble  \\ 
 \hline
      ATIS & 97.42 & \begin{tabular}{@{}c@{}} GRU, LSTM, \\ BiGRU, BiLSTM, \\ CharGRU, CharBiLSTM \end{tabular} \\
      \hline
      Banking & 91.32 & CharGRU, CharLSTM \\
      \hline
      SMP & 91.60 & \begin{tabular}{@{}c@{}} BiGRU, CharGRU, \\ CharLSTM, CharBiGRU \end{tabular}
\end{tabular}
\caption{Best RNN Ensembles}
\label{RNN Best Second-layer Ensembles}
\end{subtable}
\quad
\begin{subtable}[t]{.475\linewidth}    
\begin{tabular}{ l | c | c }
 Dataset & $F_1$ & Ensemble  \\ 
 \hline
      ATIS & 97.54 & \begin{tabular}{@{}c@{}} GRU, BiGRU, \\  BiLSTM, CNN \\ CharCNN, ABiCNN \\ ACharBiCNN \end{tabular} \\
      \hline
      Banking & 91.79 & \begin{tabular}{@{}c@{}} CharGRU, CharLSTM, \\  CharBiGRU, CharBiLSTM \\ CNN, CharCNN \\ ABiCNN, ACharBiCNN \end{tabular} \\
      \hline
      SMP & 93.55 & \begin{tabular}{@{}c@{}} CharLSTM, CharBiGRU, \\ CharBiLSTM, CharCNN, \\ ABiCNN, ACharBiCNN \end{tabular}
\end{tabular}
\caption{Best CNN/RNN Ensembles}
\label{CNN and RNN Best Second-layer Ensembles}
\end{subtable}
\caption{Second-layer Ensembles $F_1$ (\%)}
\label{Best Second-layer Ensembles}
\end{table}

We performed a series of experiments on RNN first-layer and second-layer ensembles on all three datasets. We obtained first-layer ensembles by collecting $F_1$ scores of each RNN model's three trained versions and their ensembles for ATIS, Banking, and SMP (Table \ref{RNN First-layer Ensembles}). We then performed second-layer ensemble experiments on all of the RNN models, and recored the best ensemble for each dataset in Table \ref{RNN Best Second-layer Ensembles}. 

\textbf{First-layer Ensembles}: We can derive many interesting points from these results. Firstly, each model benefits greatly from ensembling. More specifically, every first-layer ensemble performs better than at least two of its individual versions, and the vast majority outperform all three. The average $F_1$ gain for first-layer RNN ensembles is 0.66\% (ATIS), 1.50\% (Banking), and 2.66\% (SMP). 

We can also see how model complexity affects performance across different datasets. For example on ATIS, the simple GRU model is the highest scoring first-layer ensemble with 97.31\%. Adding complexity, either through bidirectionality or character embedding generally lowers performance, with CharBiGRU and CharBiLSTM getting the lowest scores of 94.62\% and 94.74\% respectively.  This is in contrast to Banking, where character embedding always helps performance (e.g. 88.15\% (GRU) to 90.02\% (CharGRU)), and bidirectionality giving mixed results.  SMP performance also contrasts with ATIS, as GRU performs worst, and character embedding helps greatly (e.g. 87.26\% (LSTM) to 91.30\% (CharLSTM)).

It is particularly interesting that Banking and SMP follow similar performance trends, despite the stark lexical and syntactic differences between English and Chinese, as well as ATIS and Banking having such different performance characteristics despite being the same language and similar size. This points to dataset/model coherence being based on subtle data characteristics. It is not obvious by inspecting ATIS and Banking data why character embedding is beneficial for Banking but detrimental for ATIS, though Banking's greater semantic diversity may be a contributing factor.

A related consideration is the relationship between bidirectionality and character embedding. In all three datasets, CharBiGRU and CharBiLSTM show significantly lower performance than their corresponding unidirectional versions, CharGRU and CharLSTM. This indicates that character embeddings and bidirectionality are somehow incompatible in an RNN architecture. This may suggest that the temporal information that is gained by reading input in both directions is lost or confused by embedding input as both words and characters. 

Another interesting result is how performance ranges between single model ensembles vary between different datasets. While Banking and ATIS only have a 2.52\% range (88.15\% GRU and BiGRU to 90.67\% CharLSTM) and 3.12\% range (94.62\% CharBiGRU to 97.74\% GRU) respectively, SMP has a 6.59\% range (84.71\% GRU to 91.30\% CharLSTM), more than twice either of the others. However, that range mostly captures the difference between character embedded models and non-character embedded models. Consequently, while character embedding may help for English datasets like Banking, it is critical for Chinese datasets. This makes intuitive sense, as Chinese characters contain rich semantic information.

\textbf{Second-layer Ensembles}: Finally, we can draw some results from the RNN second-layer ensemble experiments (Table \ref{RNN Best Second-layer Ensembles}). We can see that each best second-layer ensemble outperforms all of its single-model ensembles, showing that second-layer ensembles further increase the performance of first-layer ensembles. Each best second-layer ensemble also contains at least one GRU and at least one LSTM, again pointing to the importance of diversity in ensembling.

\subsection{CNN and RNN Ensembles}

\textbf{Second-layer Ensembles}: Table \ref{CNN and RNN Best Second-layer Ensembles} shows the best second-layer ensembles resulting from experiments with all 12 first-layer ensembles. On ATIS, multiple second-layer ensembles obtain the highest $F_1$ of 97.54\%. We show one of these in table \ref{CNN and RNN Best Second-layer Ensembles}, though other ensembles also obtain the same performance, including \{GRU, BiLSTM, ABiCNN\}. The first thing to note here is that these ensembles generally contain six or more first-layer ensembles, indicating that there is some critical mass of models needed to achieve maximum performance. Also, these ensembles have an even or very close to even split between RNN and CNN models, as well as at least one character embedded and one bidirectional model. This extends the theme discussed earlier, that greater performance is achieved through a diverse ensemble. We can see especially strong proof of this in SMP's best second-layer ensemble performance of 93.55\%. This is a nearly two percent increase from the best RNN second-layer ensemble of 91.60\%, and a more than three percent increase from the best CNN second-layer ensemble of 89.96\%. This large performance increase over two robust second-layer ensembles clearly shows that diversity of model architecture is critical for ensemble success. 

Table \ref{Best Multi-layer Ensemble $F_1$ Scores without and with Character Embeddings} shows the performance difference between the best second-layer ensembles without and with character embedding. Again we can see that character embedding is critical for the Chinese dataset, and much less so for the English datasets. Additionally, we can see that on ATIS, second-layer ensembles both with and without character embedding attain the best performance. This points to second-layer ensembling being effectively only beneficial; diverse second-layer ensembles can raise performance without any real risk of degrading performance by adding specific models, even if those models do poorly by themselves.

We can also compare performance between datasets. ATIS performs 5.75\% better than Banking, and 3.99\% better than SMP, while SMP performs 1.76\% better than Banking. Indeed, throughout both first-layer and second-layer ensembles, the same models and ensembles generally perform better on ATIS than Banking and SMP. This indicates that some datasets are generally easier for models to learn than others. The difficulty of SMP can at least in part be explained by having a much more diverse set of intents than either Banking or ATIS. Also, both Banking and SMP have a greater lexical and stylistic range than ATIS, certainly making training more difficulty. 

Finally, Table \ref{First and Second-layer Ensemble Gains} shows the average first-layer and average second-layer performance increases, as well as the total $F_1$ ensemble gain for each dataset. The average ensemble performance increases were calculated by averaging each ensemble's average difference between its performance and the performance of each of its constituent models, and the total gain was calculated as the difference between the best individual model and the best second-layer ensemble. We can see here that ensembling is much more important for some models and datasets than others, though clearly both first-layer and second-layer ensembling is more important for SMP than Banking or ATIS. The total gains represent the potential total performance increase that someone may gain through utilizing multi-layer ensembling. We see that ensembling helps less for the easier ATIS dataset, but that total increases of around 2\% and 4\% on Banking and SMP are significant. Furthermore, these total gains start from the best possible individual models; total gains from other individual models can be significantly greater. Together, these results show that ensembling can be a powerful technique for creating better intent classification systems.

\subsection{Comparison to Previous Approaches}

\begin{table}[t]
\centering
\begin{subtable}[t]{.475\linewidth}
\begin{tabular}{ l | c | c | c}
 Model & ATIS & Banking & SMP \\
 \hline
 Without Char & 97.54 & 91.04 & 91.45 \\
 With Char & 97.54 & 91.79 & 93.55 \\
 \hline
 Gain & 0.00 & 0.75 & 2.10 \\
\end{tabular}
\caption{Character embeddings gains}
\label{Best Multi-layer Ensemble $F_1$ Scores without and with Character Embeddings}
\end{subtable}
\quad
\begin{subtable}[t]{.475\linewidth}    
\begin{tabular}{ l | c | c | c }
 Model & ATIS & Banking & SMP \\ 
 \hline
 First-layer Gain & 0.54 & 1.03 & 1.91 \\
 Second-layer Gain & 0.94 & 1.24 & 2.91 \\
 Total Gain & 0.23 & 1.96 & 4.04 \\
\end{tabular}
\caption{Ensembling layer gains}
\label{First and Second-layer Ensemble Gains}
\end{subtable}
\caption{$F_1$ (\%) gains from character embedding and ensembling layers}
\label{gains from character embedding and ensembling layers}
\end{table}

\begin{table}[t]
\centering
\begin{subtable}[t]{.95\linewidth}
\begin{tabular}{ l | c | c }
 Model & $F_1$ (\%) & Development Time \\ 
 \hline
 Recursive NN \cite{Guo2014} & 95.40 & High \\
 Bidirectional GRU with Context Window \cite{Zhang2016} & 97.53 & High \\
 Attention Encoder-Decoder NN \cite{Liu2016} & \textbf{97.98} & High \\
 \hline
 Our Best Ensemble & 97.54 & Low \\
 \end{tabular}
\caption{ATIS}
\label{ATIS comparison}
\end{subtable}
\quad
\begin{subtable}[t]{.95\linewidth}    
\begin{tabular}{ l | c | c}
 Model & $F_1$ (\%) & Development Time \\ 
 \hline
Self-Inhibiting Residual CNN \cite{Lu2017} & 92.88 & High  \\
LSTM with database keyword extraction \cite{Tang2017} & \textbf{93.91} & High \\
\hline
 Our Best Ensemble & 93.55 & Low \\
 \end{tabular}
\caption{SMP}
\label{SMP comparison}
\end{subtable}
\caption{Comparison to Previous Approaches}
\label{Comparison to Previous Approaches}
\end{table}

Tables \ref{ATIS comparison} and \ref{SMP comparison} show how our best ensembles perform compared to other approaches. We can see that our best ensembles perform quite well compared to other approaches on both ATIS and SMP. Many of these models are considerably more sophisticated than our own. For example, \newcite{Liu2016} use an encoder-decoder architecture, which combines two separate neural networks, and \newcite{Tang2017} use a separate data-based keyword extraction algorithm along with a deep learning model. Our models, on the other hand, are all standard models that can be developed quickly without needing  to fine-tune the architecture. The result that our ensembles of simpler models can perform in the same range as these more sophisticated models shows that ensembling is a viable strategy for achieving very high performance on a dataset without intensive model engineering. This simplicity is also advantageous from an industry viewpoint. Since our models don't involve time-consuming development, they can more easily be deployed to a production environment.

\section{Conclusion} \label{Conclusion}

Running intent classification experiments on many variants of CNN and RNN models, as well as numerous first-layer and second-layer ensembles of those models, presented two main conclusions. First, both first-layer and second-layer ensembling can boost performance greatly, and greater diversity ensembles almost always leads to better results. Indeed, a large, diverse ensemble of straightforward models can be competitive with much more sophisticated state-of-the-art models. Consequently, researchers and engineers hoping to improve performance on intent classification should not hesitate to use multi-layer ensembling in their work.

The second main conclusion of our work concerns the idiosyncratic nature of datasets. Different models and ensembles perform quite differently on even fairly similar datasets. Although some architectures do show general performance trends across tasks and datasets, such as RNNs generally outperforming CNNs on natural language processing tasks, different datasets are idiosyncratic enough that it is extremely difficult to know \textit{a priori} which specific model will work best for a given dataset.  Network architecture, cell type, character embedding, bidirectionality, and attention can all affect datasets very differently. This idiosyncrasy strengthens the case for using ensemble methods, as they can help to distribute the insights of specific models across different datasets. We also found that diverse ensembles provide a non-negative performance gain, so that multi-layer ensembling won't hurt performance and often greatly helps it. In this way, multi-layer ensembling is a no-risk and low-effort way to improve model performance on intent classification. 

These conclusions support our hypothesis that multi-layer ensembling can provide significant performance increases compared to their constituent models. In this work we also saw that multi-layer ensembling of simple models can compete with more complex models, and that ensembling provides a no-risk method to improve performance on intent classification across different languages and domains. 

\bibliography{cloudminds}
\bibliographystyle{acl}

\end{document}